\documentclass[10pt,twocolumn,letterpaper]{article}

\usepackage[pagenumbers]{cvpr} % To force page numbers, e.g. for an arXiv version

\definecolor{cvprblue}{rgb}{0.21,0.49,0.74}
\usepackage[pagebackref,breaklinks,colorlinks,allcolors=cvprblue]{hyperref}
\newcommand{\cmark}{\ding{51}}%
\newcommand{\xmark}{\ding{55}}%
\usepackage{physics, amsmath, bm}
\usepackage{microtype}
\usepackage{graphicx}
\usepackage{booktabs} % for professional tables
\usepackage{multirow}
\usepackage[table]{xcolor}

\usepackage{amsmath}
\usepackage{dsfont}
\usepackage{float}
\usepackage{multirow}
\usepackage{multicol}
\usepackage{tablefootnote}
\usepackage{pdflscape}
\usepackage{pifont}
\usepackage{subcaption} %  for subfigures environments 
\usepackage{booktabs}

\def\paperID{17} % *** Enter the Paper ID here
\def\confName{CVPR}
\def\confYear{2026}

\title{MEBench: A Novel Benchmark for Understanding Mutual Exclusivity Bias in Vision-Language Models}

\author{Anh Thai$^1$ \quad Stefan Stojanov$^{1}$\quad Zixuan Huang$^{2}$
 \quad Bikram Boote$^{2}$ \quad James M. Rehg$^{2}$\\
$^1$Georgia Institute of Technology,
$^2$University of Illinois, Urbana-Champaign}

\begin{document}
\twocolumn[{%
\renewcommand\twocolumn[1][]{#1}%
\maketitle

\begin{center}
    \centering
    \captionsetup{type=figure}
    \vspace{-3ex}
    % \fbox{\rule{0pt}{3in} \rule{\linewidth}{0pt}}
    \includegraphics[width=\textwidth]{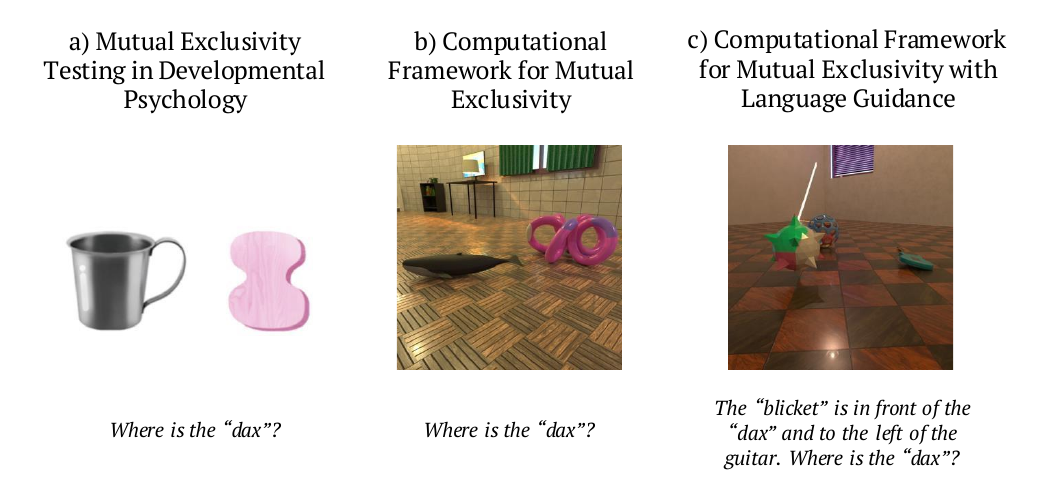}
    \vspace{-3.6ex}
    \caption{\textbf{Mutual Exclusivity Bias Evaluation Settings.} (a) Traditional ME bias evaluation in developmental psychology and early computational studies~\cite{gandhi2020mutual}, (b) MEBench setup for classic ME bias testing, and (c) MEBench setup for evaluating ME bias in conjunction with spatial reasoning.}
\label{fig:problem_intro}
    
\end{center}%
}]
\begin{abstract}
This paper introduces MEBench, a novel benchmark for evaluating mutual exclusivity (ME) bias, a cognitive phenomenon observed in children during word learning. Unlike traditional ME tasks, MEBench further incorporates spatial reasoning to create more challenging and realistic evaluation settings. To facilitate controlled experimentation, we also present a flexible and scalable data generation pipeline that supports the construction of diverse annotated scenes. We assess the performance of various vision-language models (VLMs) on this benchmark using novel evaluation metrics that capture key aspects of ME-based reasoning. We find that these VLMs exhibit weak ME bias, while showing some ability to leverage extra spatial context to resolve ambiguity in multiple novel object settings. Project page: \url{http://mebench.github.io/}.
\end{abstract}    

\section{Introduction}

During a developmental phase known as the vocabulary spurt~\cite{mcmurray2007defusing}, toddlers remarkably acquire 10–20 new words per week. This rapid word learning is supported by powerful inductive biases that help children resolve the ambiguity of mapping spoken words to real-world referents, referred to as object-label mapping. One such bias, mutual exclusivity, is the assumption that each object has a unique label. When presented with a mix of familiar and unfamiliar objects, children tend to associate novel words with the unknown items. For instance, if a caregiver says, “Look at the blicket,” while the child sees her favorite Ducky, a Mickey Mouse figure, and a novel toy, mutual exclusivity guides the child to infer that “blicket” refers to the toy whose name is unknown.

Inspired by such cognitive mechanisms, researchers in computer vision have begun exploring how inductive biases can be incorporated into learning models to improve generalization and interpretability. Among these, shape bias, the tendency to associate category membership with object shape, has received significant attention~\cite{stojanov2021using,padmanabhan2023lsfsl,prasad2022implicit}, particularly due to its relevance in 3D object categorization. In contrast, mutual exclusivity bias, despite being a foundational aspect of human language learning~\cite{gandhi2020mutual,markman1988children,markman2003use}, remains relatively underexplored in computational settings.

With the rise of Vision-Language Models (VLMs) that aim to mirror human-like reasoning by grounding language in visual inputs, there is now an opportunity to study mutual exclusivity within a computational framework. These models provide a natural platform for investigating whether and how such biases emerge, especially in complex visual scenes that require reasoning.

In this work, we introduce MEBench, a new benchmark for evaluating mutual exclusivity bias in VLMs. Inspired by classic studies in developmental psychology, MEBench formulates the problem as an inference task: given an image containing both known and novel objects, along with a novel label, the model must (1) identify and localize the known categories and (2) associate the novel label with the correct unknown instance by leveraging mutual exclusivity. MEBench goes beyond prior work that focuses solely on object-label association~\cite{thai2023low} by introducing realistic, cluttered scenes and additional reasoning requirements. Specifically, models must not only detect and differentiate between known and novel objects but also apply spatial reasoning and contextual cues to resolve ambiguity when multiple unfamiliar items are present.

Since no real-world dataset exists for this task, we develop a scalable synthetic data generation pipeline. This pipeline transforms any 3D dataset of object categories into realistic scenes and includes a curated set of procedurally generated novel objects, using Blender’s geometry nodes~\cite{blender}. To prevent lexical leakage, novel object labels are sampled from synthetic, non-English-like words (e.g., “dax,” “toma”) commonly used in human studies of mutual exclusivity~\cite{smith2008infants}.

We systematically evaluate state-of-the-art VLMs on MEBench and analyze their performance across multiple subtasks, revealing both capabilities and limitations in their reasoning behavior. Mastery of MEBench would mark a meaningful step toward equipping AI systems with robust zero-shot generalization capabilities, essential for adaptive applications in real-world domains such as home robotics and human-AI interaction.

In summary, our contributions are:
\begin{enumerate}
\item We introduce MEBench, a computational framework for evaluating mutual exclusivity (ME) bias, and extend its evaluation beyond traditional settings by incorporating spatial reasoning with contextual scene descriptions.
\item We develop a flexible synthetic data generation pipeline that enables controlled experimentation and systematic analysis of factors affecting ME-based learning.
\item We benchmark a diverse set of vision-language models (VLMs) on MEBench, and find that modest increases in scene complexity cause a cliff in ME performance for all models, and leveraging spatial context improves performance in ambiguous multi-novel settings.
% while these models exhibit weak ME bias, they demonstrate positive spatial reasoning capabilities.
\end{enumerate}
\section{Related Work}

% Because we are the first to provide a computational framing of mutual exclusivity (ME), there is no direct prior work to compare to.~\cite{gandhi2020mutual} demonstrated that existing deep models fail at ME, but did not provide a comprehensive framing or SOTA methods. We now review three bodies of related work.

\textbf{Computational Mutual Exclusivity Modeling.} The work most closely related to ours is~\cite{thai2023low}, which introduces a generalized low-shot object learning framework that requires applying the mutual exclusivity assumption to associate a novel label with the correct novel object. While our approach also relies on mutual exclusivity, it differs in two key aspects. First, we incorporate a curated set of procedurally generated novel objects paired with pseudo words as labels, minimizing lexical leakage during evaluation. Second, we extend the task beyond traditional mutual exclusivity by introducing a setting with multiple novel objects, where additional reasoning, such as interpreting spatial cues, is required to resolve ambiguity. This level of disambiguation is not addressed in~\cite{thai2023low}. Another prior work, ~\cite{gandhi2020mutual} demonstrated that existing feed-forward deep models fail at ME, but did not provide a comprehensive framing or benchmarking SOTA methods.

Other recent works have investigated various learning scenarios to explore the strategies children employ when acquiring new concepts~\cite{jiang2023mewl,pmlr-v161-agrawal21a,hill2020grounded}. We complement these efforts by introducing a comprehensive benchmark that studies the mutual exclusivity bias commonly observed in infants during the initial stages of word learning.

\textbf{Language-guided Object Detection/Segmentation Tasks.} Open-category object detection and segmentation~\cite{xu2022odise,joseph2021towards} settings require learning models to detect and segment both known and novel objects---novel to the localization models, though not necessarily to the LLMs---by prompting the LLMs. In contrast, our task focuses on learning object-label mappings, where labels may fall outside typical pretraining vocabularies, such as pseudo-names like “dax” commonly used in psychology studies~\cite{smith2008infants}.

Referring and reasoning segmentation~\cite{lai2024lisa,yuan2025sa2va,rasheed2024glamm} tasks require models to segment objects in an image based either on explicit natural language attribute-based descriptions or functional reasoning. Our setting extends beyond these tasks by requiring models to leverage ME bias to associate a novel label (a pseudo-name unfamiliar to the LLMs) with the novel object. When multiple novel objects are present, the model must further reason about spatial relationships to resolve ambiguity. Our approach emphasizes inter-object relationships, requiring the model to develop a holistic scene understanding alongside object-centric reasoning. 
% \subsection{Open-world Learning} The objective of open-world learning methods~\cite{} is to correctly identify objects from the known data distribution, and further distinguish if data comes from a novel distribution. Unlike their closed-set counterparts, these methods are designed to accommodate both known and unknown data during evaluation. In this setting, labeling new data is not considered, and such instances are typically assigned the label of "unknown." Recent works~\cite{} have explored learning in an incremental manner, continually updating the known data distribution to adapt to new information. In our proposed task, we require the models to perform open-world learning as a part of the solution.

% \textbf{Computational Frameworks Motivated by Developmental Psychology.} Recent works have  investigated various learning scenarios to explore the strategies children employ when acquiring new concepts~\cite{jiang2023mewl,pmlr-v161-agrawal21a,hill2020grounded}. We complement these efforts by introducing a comprehensive benchmark that studies the mutual exclusivity bias commonly observed in infants during the initial stages of word learning. Because we are the first to provide a computational framing of mutual exclusivity (ME), there is no direct prior work to compare to.~\cite{gandhi2020mutual} demonstrated that existing deep models fail at ME, but did not provide a comprehensive framing or SOTA methods. 

\textbf{Synthetic 3D Datasets and Generators.} 
% Synthetic 3D datasets have been widely used to study various tasks~\cite{stojanov2021using,stojanov2022learning,weinzaepfelcroco,greff2022kubric,savva2019habitat} where large-scale real-world data is unavailable. 
Advances in realistic rendering engines~\cite{greff2022kubric, gan2021threedworld, savva2019habitat} and image generation models~\cite{rombach2022high,saharia2022photorealistic} have helped narrow the sim-to-real gap, improving generalization to real-world scenarios, and have been widely employed to study various tasks~\cite{stojanov2021using,stojanov2022learning,weinzaepfelcroco,greff2022kubric,savva2019habitat} where large-scale real-world data is unavailable. In our work, we use synthetic data to address a novel problem where real-world datasets do not yet exist. This approach enables the creation of diverse, realistic, and controllable environments for scalable experimentation.

\textbf{Vision-Language Detection/Segmentation Models.} With advancements in large vision-language models (VLMs), the ability of AI systems to comprehend images and respond to language prompts related to visual content has improved significantly~\cite{openai2023gpt4v,geminiteam2024geminifamilyhighlycapable,anthropic2024claude3,lu2024deepseekvl,li2024llava} 
% have demonstrated remarkable capabilities in understanding the visual world through images, enabling them to analyze and respond to complex questions about image content. 
% These models have been further extended to video processing and understanding~\cite{kalarani2024unveiling,chen2023videollm}, 3D scene comprehension~\cite{hong20233d,huang2023embodied,wang2023chat}, and various other complex reasoning tasks~\cite{cheng2024empowering,zhang2023planning}. 
% They are typically trained on large-scale datasets, enhancing their generalization ability across diverse domains and enabling them to perform well even on data distributions beyond those encountered during training.
Beyond general image understanding, other works~\cite{wang2023cogvlm,yuan2025sa2va,zhang2025omg,rasheed2024glamm,wu2024f} focus on object grounding, which involves localizing and reasoning about objects within an image and outputting their locations, rather than solely providing text-based responses. These models require a precise scene representation to accurately ground objects based on language prompts. In this work, we use these models as our baselines for our MEBench due to their strong object detection capabilities and their ability to reason effectively with language inputs.
%%%% ABC, Toys, ShapeNet datasets
\begin{figure*}[t!]
\centering
\includegraphics[width=\linewidth]{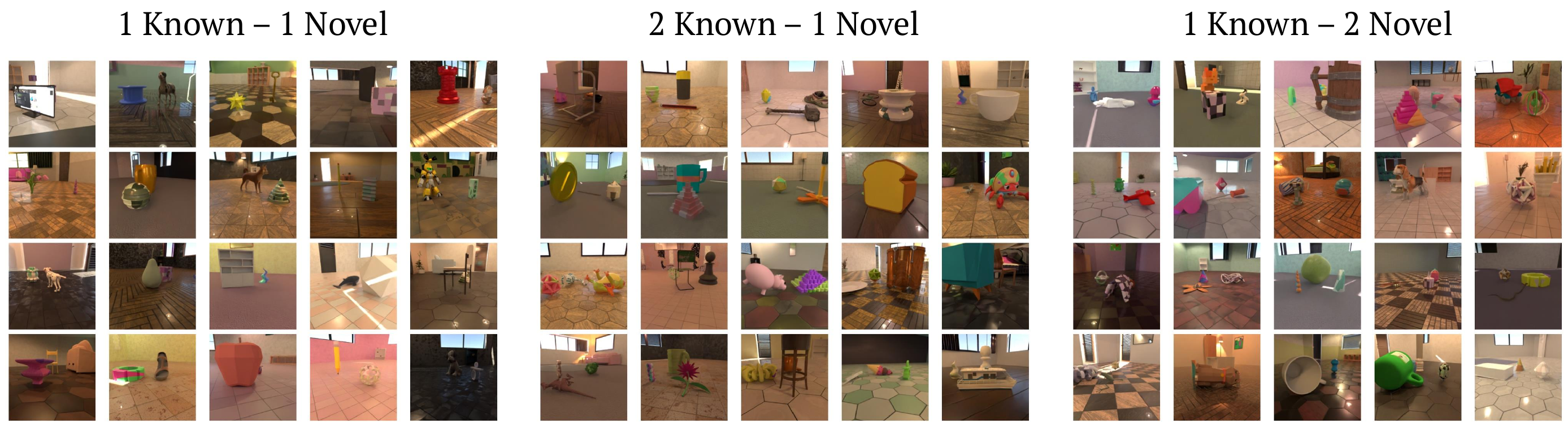}
\caption{\textbf{Example of Rendered Data for the MEBench Benchmark.} We systematically generate diverse object configurations within varied room backgrounds, ensuring photorealistic renderings that capture realistic spatial arrangements and lighting conditions.}
\label{fig:example_rendering_new}
\vspace{-5pt}
\end{figure*}
% \caption{}
% \label{fig:setting_overview}
% \vspace{-20pt}

\section{MEBench}
In this section, we first formulate the task in Section~\ref{sec:problem_formulation}, followed by a detailed description of the data generation pipeline in Section~\ref{sec:data_generation}. Finally, we introduce the various data variants used in our study in Section~\ref{sec:data_variants}.
\subsection{Task Formalization}\label{sec:problem_formulation}
Given an RGB image containing multiple objects—some from known everyday categories and one or more novel objects—the model is asked to localize the referent of a novel label (e.g., “Where is the blicket?”). The task involves three subtasks: (1) \textbf{Object Detection}, further divided into: \textit{Object Localization}---Identifying and localizing all objects present in the scene; and \textit{Open-world Recognition}--- Differentiating between known and novel instances using visual information; (2) \textbf{Novel Label Assignment}---Applying the mutual exclusivity assumption to correctly associate the novel label with the novel object.

When multiple novel objects are present, mutual exclusivity alone is insufficient to resolve ambiguity. To address this, we introduce spatial descriptions (e.g., “The dog is to the right of the blicket”) that encode inter-object relations. The model must then perform: (3) \textbf{Spatial Reasoning}---the ability to interpret spatial descriptions in text to facilitate accurate novel label assignment. Note that this additional input can also be incorporated into the single unknown object scenario.

\subsection{Data Generation Pipeline}\label{sec:data_generation} 
% In this subsection, we describe our data generation pipeline. An overview can be seen in~\ref{fig:data_generation_pipeline}. 
Our system is designed to be compatible with any dataset of 3D object categories. For our benchmark, we use the Toys4K~\cite{stojanov2021using} dataset, chosen for its large number of categories (105) and diverse toy-like object appearances, which closely resemble the variability encountered in real-world child learning scenarios. This dataset comprises common, everyday object categories (e.g., "car", "dog"), which serve as the known categories in our benchmark.

\begin{figure*}[t!]
\centering
\includegraphics[width=0.8\linewidth]{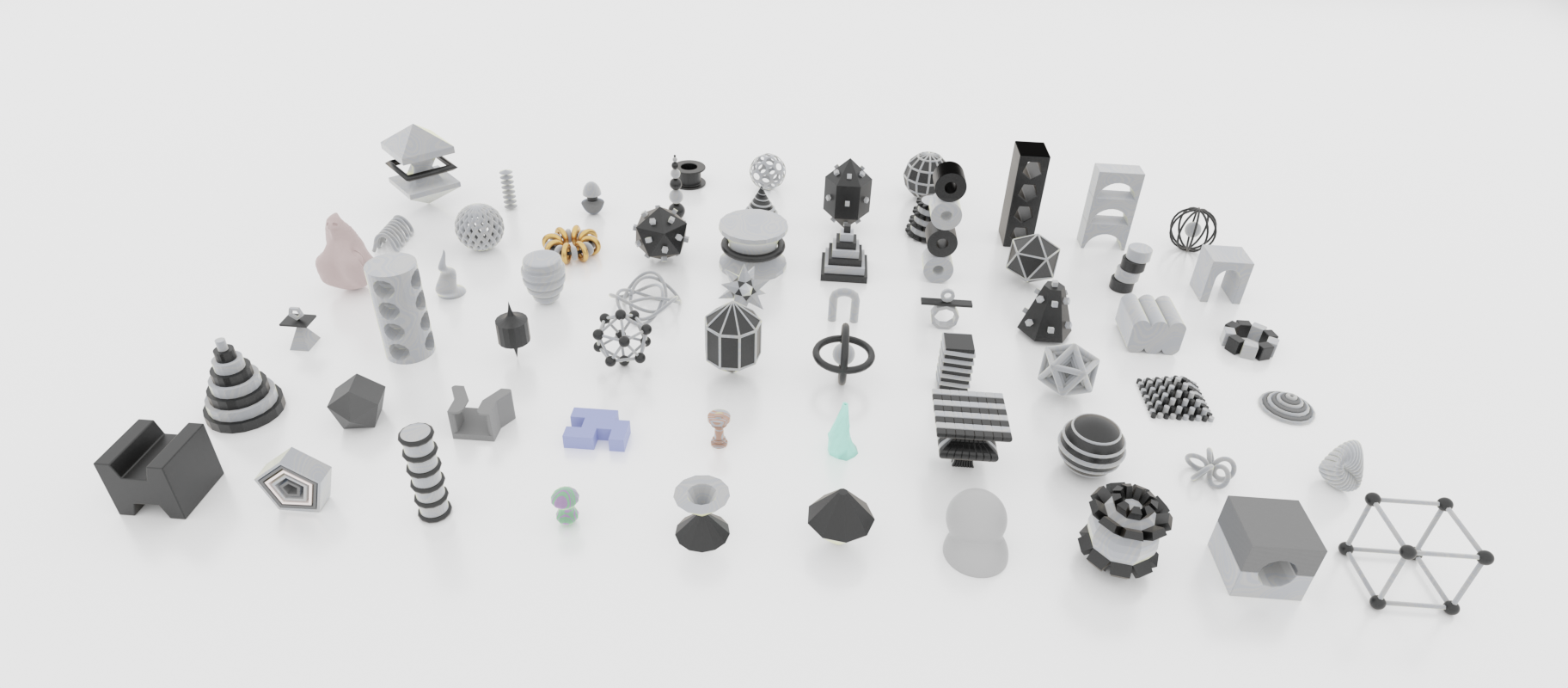}
\caption{\textbf{Novel Objects in MEBench.} To minimize lexical leakage during evaluation, we constructed a database of novel objects using procedural generation in Blender~\cite{blender} with geometry nodes from GeoShapeV2~\cite{geoshapes} and Thingi10K~\cite{zhou2016thingi10k}, paired with pseudo-words as labels.}
\label{fig:novel-obj}
\vspace{-15pt}
\end{figure*}

    \noindent\textbf{Novel Objects.} We introduce a curated set of 64 novel objects (Figure~\ref{fig:novel-obj}), manually designed and procedurally generated using geometric nodes in Blender~\cite{blender}, ensuring unique and diverse structures. The scale of our novel object set is comparable to prior benchmarks that evaluate generalization to unseen or uncommon categories (e.g., Bogard-HOI~\cite{jiang2022bongard} considers 16–64 unseen classes, and other work~\cite{pi2024uouo} benchmarks VLMs on 50 uncommon objects), while the developmental psychology NOUN database ~\cite{horst2016novel} contains 45–64 objects with unusual names. This set is specifically crafted to minimize lexical leakage when paired with pseudo-word labels: foundation models trained on web-scale data have likely learned associations between common objects and their names, but are unlikely to have learned stable name–visual mappings for these synthetic, abstract, non-semantic shapes. To further ensure fairness, we randomize the pairing between novel objects and pseudo-word labels at inference time, preventing any fixed object–word assignment from being exploited.
    % \ssnote{I would find some words to describe the properties of these objects, yes they have unique and diverse structures, but what makes them good for this purpose?}
% By using synthetic, previously unseen objects, we create a more controlled evaluation setting for assessing zero-shot generalization and mutual exclusivity bias in models. During rendering, these novel objects are assigned randomized materials and colors, ensuring diversity in appearance and preventing models from relying on any texture-based shortcuts for recognition.

\noindent\textbf{Background.}
% We generate room backgrounds containing typical distraction objects commonly found in specific environments, such as living rooms and bedrooms.
% two types of backgrounds based on the presence of distraction objects in the scene. When distraction objects are present, we use a room background. Alternatively in the case where the background distractions do not exist, we apply a blurred HDRI background and place the objects on a plane with a floor-like material. This helps minimize potential distractions for the models, such as windows or complex wall textures.
% \noindent\underline{\textit{Room Generation.}} 
To create diverse and realistic backgrounds, we generate room environments as the backdrop for our scenes. These rooms are primarily living rooms and bedrooms, as they represent the most natural settings for child play. Our room generation is based on Infinigen~\cite{infinigen2024indoors}, with each room including varied background object configurations. Lighting conditions are naturally derived from indoor sources or outdoor light filtering through open doors and windows (see Figure~\ref{fig:example_rendering_new}).

\noindent\textbf{Data Rendering.} During each scene rendering, we first randomly select a subset of objects from the known categories and a set of novel objects from our curated collection. The total number of objects and the number of novel instances are determined based on user-defined input parameters. To achieve realistic object placement, we use rigid body simulation to generate natural rotational poses. Objects are then scaled and positioned at random locations within the scene while ensuring that no two objects collide. 
% Further, we prevent intersections between placed objects and background elements. To enhance diversity and naturalistic scene appearance, we sample multiple camera viewpoints for each scene, capturing variations in perspective, depth, and occlusions.

%%%% ABC, Toys, ShapeNet datasets
% \caption{}
% \label{fig:setting_overview}
% \vspace{-20pt}

\noindent\textbf{Scene Description Generation.} We generate scene descriptions as additional contextual inputs for each view of the scene, rather than for the entire 3D scene. This is because some spatial relationships in 3D are inherently viewpoint-dependent terms like ``left," ``right," ``in front of," and ``behind" can vary depending on the observer’s perspective. For each view, we first construct a scene graph using object bounding boxes and the corresponding depth map. In this graph, objects are represented as nodes, while pairwise spatial relationships form the edges. We then translate the structured scene graph representation into plain English descriptions. For instance, given the scene graph expression: ``Dog":\{``to left of":[``dax", ``pig"]\} we generate the natural language description: ``The dog is to the left of the dax and the pig." Notably, this scene description generation process is fully deterministic, ensuring that all pairwise object relationships are consistently included in the description. This eliminates potential ambiguities and provides a structured yet flexible input format for downstream reasoning tasks.
% For instance, given the scene graph expression: ``Dog":\{``to left of":[``dax", ``pig"]\} we generate the natural language description: ``The dog is to the left of the dax and the pig." 
% Notably, this scene description generation process is fully deterministic. This eliminates potential ambiguities and provides a structured yet flexible input format for downstream reasoning tasks.

% We generate the 3D scene before creating scene descriptions for each view because spatial relationships in 3D can be ambiguous. For example, directional terms such as behind, in front of, left, and right depend on the viewpoint and cannot be directly described in 3D space but can be precisely defined in 2D images. Therefore, we first generate the 3D scene before generating scene graph and translate it into English descriptions for each specific view.

\noindent\textbf{View-point Selection for Inference.} During inference, we select rendered viewpoints where all objects in the scene are visible. An object is considered visible in a given view if its segmentation mask occupies at least 200 pixels in a $224\times 224$ resolution image. To further minimize variability in object visibility, we run each model three times, each with a different set of selected viewpoints, and a different novel object–pseudo-word assignment.

% This criterion ensures that: 1) All models are evaluated fairly, eliminating potential viewpoint bias where certain views contain fewer visible objects than others; 2) The designated experimental setting is strictly maintained, for instance, in the 2K-1U setting, all three objects (two known, one unknown) are always visible during inference. To further minimize variability in object visibility, we run each model three times, each with a different set of selected viewpoints.
\begin{table}

\caption{\textbf{Overview of Data Variants for Evaluating VLM Baselines.} The dataset variants are categorized into three groups, each designed to assess a specific subtask: object localization, mutual exclusivity (ME) bias, and spatial reasoning ability.}
\begin{center}
% \caption{Characteristics of different single-view 3D shape reconstruction methods. }

\begingroup
\setlength{\tabcolsep}{3pt} % Default value: 6pt
\renewcommand{\arraystretch}{1.2} % Default value: 1

\vspace{-10pt}
\scalebox{0.7}{

% \begin{table*}[h]
% \centering
% \caption{PartNetE}
 \begin{tabular}{l|c c|c c c}
        \toprule
        \multirow{3}{*}{Setting}& \multicolumn{2}{c|}{Data Axis} & \multicolumn{3}{c}{Model Assessment}\\
        % \cline(2-5)
        \cmidrule{2-6}
        % & #Known Obj. & #Novel Obj. & Distraction Obj. & Obj. Localizsation & ME Bias & Spatial Reasoning\\

          & \begin{tabular}{@{}c@{}}\#Known \vspace{-3pt}\\ Obj. \end{tabular} & \begin{tabular}{@{}c@{}}\#Novel \vspace{-3pt}\\ Obj. \end{tabular} & \begin{tabular}{@{}c@{}}Obj. \vspace{-3pt}\\ Localization \end{tabular} & \begin{tabular}{@{}c@{}}ME \vspace{-3pt}\\ Bias \end{tabular} & \begin{tabular}{@{}c@{}}Spatial \vspace{-3pt}\\ Reasoning \end{tabular}\\
        \midrule
       1K-0U & 1 & 0  & \cmark & \xmark & \xmark \\
       \midrule
       1K-1U & 1 & 1  & \cmark & \cmark & \xmark \\
       2K-1U & 2& 1 & \cmark & \cmark & \xmark \\
       \midrule
       1K-2U & 1 & 2  & \cmark & \cmark & \cmark\\
       \bottomrule

\end{tabular}

}
% \caption{.}
\endgroup
 \label{tbl:data_setting_new_chap6}

\end{center}
\end{table}

\subsection{Data Variants}\label{sec:data_variants} 
To enable a comprehensive analysis and gain deeper insights into the limitations of different methods, we generate datasets with progressively increasing levels of difficulty. Each variant is specifically designed to evaluate three key model capabilities:
(1) Object Localization, (2) Mutual Exclusivity (ME) Bias, and (3) Spatial Reasoning. These capabilities are assessed along two primary axes: Number of known objects and Number of novel objects. To maintain clarity, we define data variant acronyms based on these axes. For example, 1K-0U represents a scenario with 1 known object and 0 novel objects. These data variants provide a framework for systematically testing model capabilities, distinguishing basic object detection from higher-level reasoning tasks such as ME bias and spatial inference (Table~\ref{tbl:data_setting_new_chap6}).
% This structured dataset design ensures a controlled evaluation of how different factors impact model performance.

\noindent\textbf{Object Localization: Evaluating Detection Ability.} The first group consists of only one known object, without any novel objects: 1K-0U. This setting assesses the model’s ability to localize known objects. A model that performs well in this group demonstrates object detection capabilities, independent of novel object inference.

\noindent\textbf{Mutual Exclusivity Bias: Assigning Novel Labels Correctly.} The second group introduces one novel object alongside known objects (1K-1U and 2K-1U). This group evaluates the model's ability to apply the ME assumption, which assigns the novel label to the novel object. 
% Additionally, comparing model performance on 1K-1U-YesD versus 2K-1U-YesD provides insight into the model’s object detection robustness as the number of known objects increases.

\noindent\textbf{Spatial Reasoning: Disambiguating Multiple Novel Objects.} The third group contains two novel objects: 1K-2U, introducing inherent ambiguity that cannot be resolved solely through ME bias. This setting evaluates the model’s ability to leverage spatial descriptions to disambiguate novel objects based on their relationships within the scene.

Due to the procedural nature of our data generation pipeline, the benchmark can be readily scaled to more challenging settings, such as scenes with additional known and novel objects or more cluttered backgrounds. In this work, however, we focus on systematically analyzing model behavior across the core evaluation axes above, rather than substantially increasing scene complexity.

% \noindent\textbf{Mutual Exclusivity as Bias vs. Pragmatic Reasoning.} In the developmental psychology community, there has been a long-standing debate surrounding the phenomenon of mutual exclusivity. Some researchers argue that it reflects an inherent bias, where children rely on language-specific knowledge, specifically, the principle that objects typically have only one label, to make inferences~\cite{mcmurray2012word,horst2008fast}. Others contend that mutual exclusivity arises from more general social-pragmatic principles, wherein children use the speaker’s intent to guide their inferences~\cite{frank2009using,bohn2019pervasive}.

% In this work, we do not take a position on either side of this debate. Instead, we propose a computational framework for evaluating mutual exclusivity behavior in AI models, specifically vision-language models (VLMs), in the context of novel word–object association. We draw inspiration from both perspectives: data variants that assess model performance in minimal contexts (e.g., ``Where is the dax?”) are aligned with the bias-driven interpretation, while those that provide additional contextual or relational information that require the model to reason about the scene, more closely reflect the pragmatic view of mutual exclusivity.

% \section{Vision-Language Models Baselines}

\section{Experiments}

\subsection{Baselines}

We evaluate our MEBench benchmark on SOTA VLM baselines, consisting of both closed-source and open-source models: CogVLM~\cite{wang2023cogvlm}, Gemini 2.0~\cite{geminiteam2024geminifamilyhighlycapable}, Sa2VA~\cite{yuan2025sa2va}, OMG-LLaVA~\cite{zhang2025omg}, LISA~\cite{lai2024lisa}, LLaVA-OV~\cite{li2024llava}, and F-LMM~\cite{wu2024f}. 
% These baselines are categorized into three primary model types based on their output modalities: (1) Text-only output, (2) Text + Bounding Box output, and (3) Text + Segmentation Mask output (see~\ref{fig:vlm_arch}). 
% Each of these models has been pre-trained and fine-tuned on large-scale datasets, achieving SOTA performance across various vision-language tasks and benchmarks. For a comprehensive summary of these baselines, please refer to~\ref{tbl:vlm_baseline}.
Most of these VLMs are trained primarily for text-guided object grounding, such as detecting or segmenting objects based on language prompts, which makes them particularly relevant baselines for our task given its close connection to object detection. In contrast, LLaVA-OV~\cite{li2024llava} is designed as a more general VQA and reasoning model. The diversity in model architectures and training objectives allows us to conduct a comprehensive analysis, gaining deeper insights into how different VLMs perform on our benchmark across various reasoning and perception challenges. For all experiments, we evaluate each baseline three times, using a different subset of scene viewpoints and a different novel object–pseudo-word assignment in each run. We observe that performance is highly consistent across runs, with negligible standard deviation for all models (Figure~\ref{fig:combine_detection}).

\subsection{Data}

% \input{tables_tex/chapter6/data_variants}

% \subsection{Data Variants}

% As discussed in~\ref{chapter4}, we evaluate our baselines on five data variants, each varying along one or more of the three key ME axes: the number of known objects, the number of novel objects, and the presence of distraction objects in the background. Each variant assesses specific subtasks, including object localization, ME bias, and spatial reasoning (see~\ref{tbl:data_setting_new_chap6}). These variants are categorized into three groups (see~\ref{sec:data_variants}): 1K-0U-NoD and 1K-0U-YesD assess object detection ability, 1K-1U-YesD and 2K-1U-YesD additionally evaluate ME bias, and 1K-2U-YesD measures spatial reasoning. For details about how data is constructed and rendered, please see~\ref{sec:data_generation}.

% \input{figure_tex/data_generation}

Using the data generation pipeline from Section~\ref{sec:data_generation}, we generate 100 scenes per setting. This number provides substantial variation in object combinations, configurations, backgrounds, and lighting conditions, while keeping the dataset manageable for evaluation. Additionally, each scene is rendered from 25 randomly sampled viewpoints, ensuring diverse viewing perspectives. In total we achieve 2,500 images per setting, which provides robust coverage of object appearances, spatial arrangements and viewpoints.
% In total we achieve 2,500 images per setting, which provides robust coverage of object appearances, spatial arrangements and viewpoints. 
During inference, we ensure that all objects are visible in the selected views to maintain consistency and fairness in evaluation (see Section~\ref{sec:data_generation}).

\subsection{Evaluation Protocol \& Metrics}\label{sec:chap6_metric}

In this subsection, we outline the evaluation criteria for each subtask and the metrics used to assess model performance.

\noindent\textbf{Object Localization.} We assess the models' ability to detect objects using the standard Average Precision at IoU threshold ($AP@t$) metric, commonly used in object detection tasks~\cite{redmon2016you,wang2025yolov10}.
% This metric evaluates model performance across different IoU thresholds by measuring the alignment between predicted and ground-truth bounding boxes. 

% While our primary focus is on object localization performance for known objects, following the conventional evaluation for object detection, we also assess performance on novel objects under the assumption that the model possesses an understanding of objections, i.e., the ability to detect all objects present in the scene. Errors in detecting novel objects may indicate two potential weaknesses: (1) a lack of mutual exclusivity (ME) bias, preventing the model from correctly associating novel labels with novel objects, or (2) an inability to visually parse the scene, leading to failures in recognizing and localizing objects altogether. This assumption holds across all the sub-tasks evaluated in our study.

\noindent\textbf{Mutual Exclusivity Bias.} Mutual exclusivity (ME) bias refers to the tendency to assign a novel label to a novel object, rather than to an already familiar one. To assess whether a model exhibits ME bias, we evaluate its ability to assign the pseudo-word label to the correct novel object, conditioned on successfully recognizing all familiar objects present in the scene. This conditioning is important because it excludes cases where the model fails to recognize a known object and therefore cannot meaningfully apply mutual exclusivity. 
We first define $p(x_{n \rightarrow n})$ as the probability of correctly assigning the novel label to a novel object, equivalent to the ME score from~\cite{gandhi2020mutual}, and conceptually similar to $AP@0.5$ metric for novel objects. 
% the $AP@t$ metric for novel objects and 
Similarly, $p(x_{k \rightarrow k})$ denotes correct assignment of known labels, and $p(x_{n \rightarrow k})$ captures incorrect assignment of the novel label to a known object.
\begin{figure*}[t]
% \vspace{-10pt}
\begin{subfigure}{0.33\textwidth}
\centering
	\includegraphics[width=\linewidth]{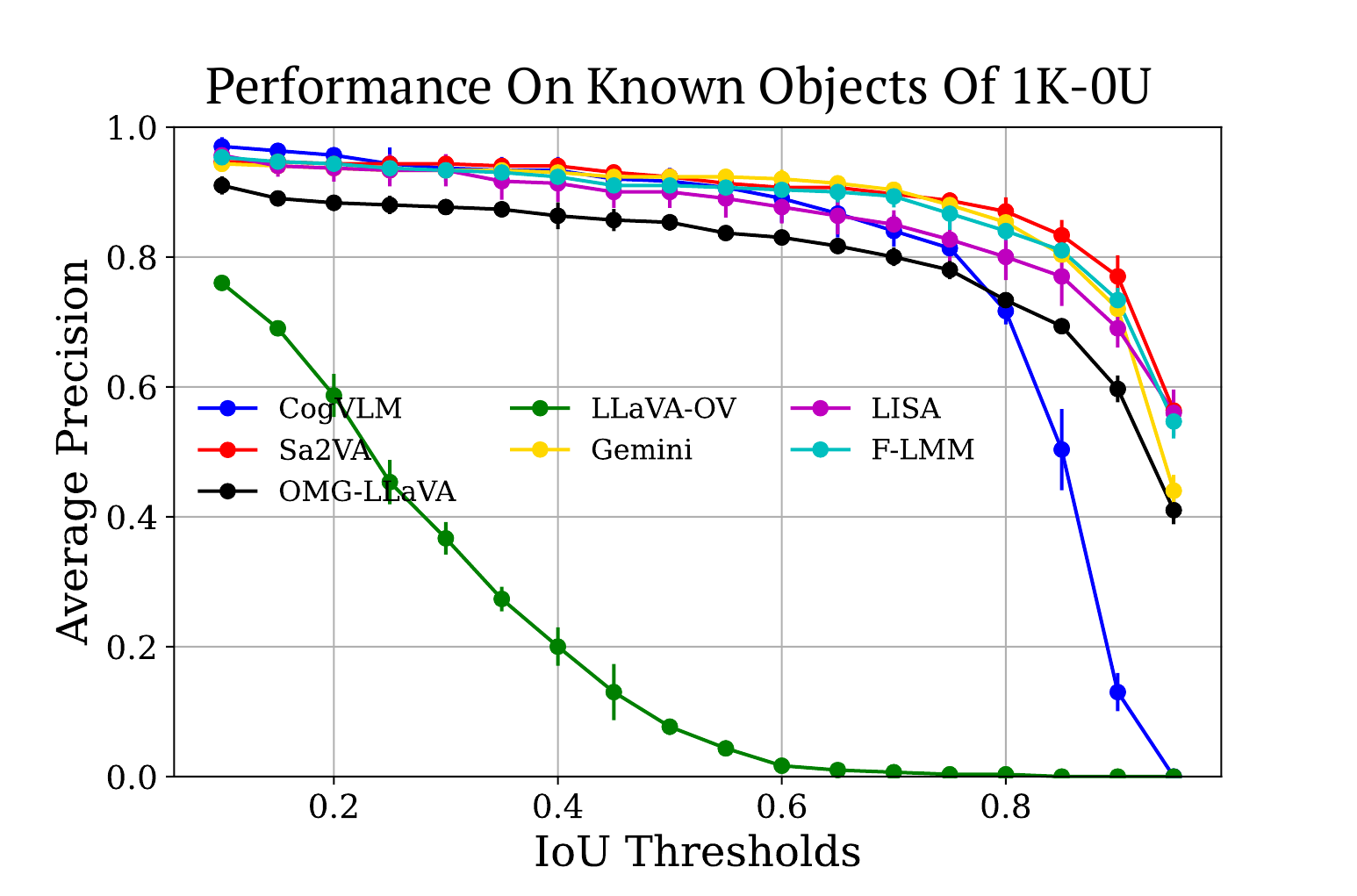}
	% \caption{\textbf{Non-overlapping 2D Mask Proposal.} We address the issue of overlapping masks produced by SAM. The masks are first sorted by their areas. Subsequently, the smaller masks are stacked on top of the larger ones. Non-overlapping masks are obtained by taking the visible segment of each mask.}

  % \vspace{-5pt}
   \caption{}
	\label{fig:1K0U}
   
\end{subfigure}
  \vspace{-5pt}
\begin{subfigure}{0.33\textwidth}
\centering
	\includegraphics[width=\linewidth]{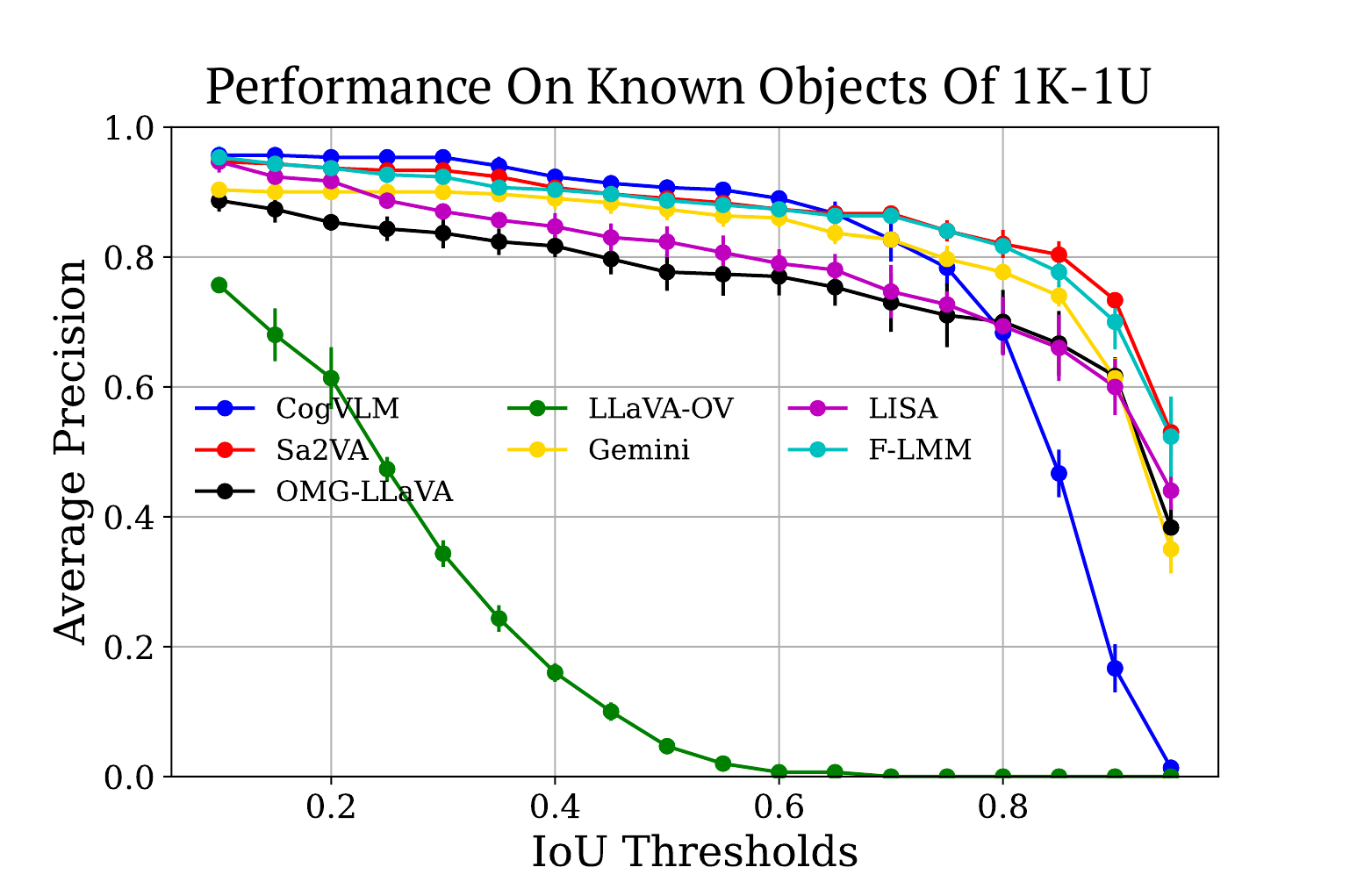}
	% \caption{\textbf{Non-overlapping 2D Mask Proposal.} We address the issue of overlapping masks produced by SAM. The masks are first sorted by their areas. Subsequently, the smaller masks are stacked on top of the larger ones. Non-overlapping masks are obtained by taking the visible segment of each mask.}

  % \vspace{-5pt}
   \caption{}
	\label{fig:1K1U}
   
\end{subfigure}
  \vspace{-5pt}
\begin{subfigure}{0.33\textwidth}
\centering
	\includegraphics[width=\linewidth]{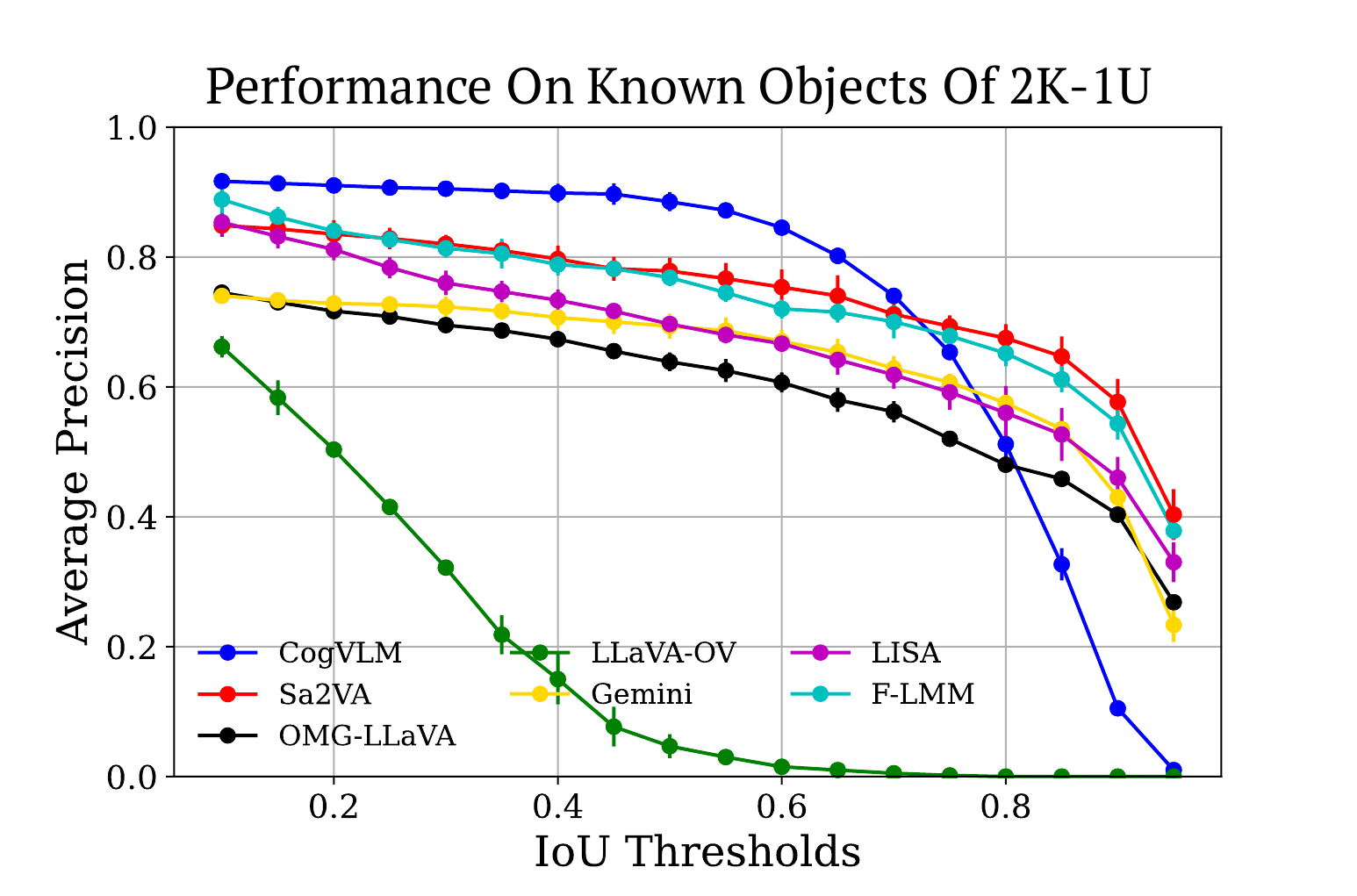}
	% \caption{\textbf{Non-overlapping 2D Mask Proposal.} We address the issue of overlapping masks produced by SAM. The masks are first sorted by their areas. Subsequently, the smaller masks are stacked on top of the larger ones. Non-overlapping masks are obtained by taking the visible segment of each mask.}

	% \label{fig:2K1U}
  % \vspace{-5pt}
   \caption{}
   \label{fig:2K1U}
\end{subfigure}
  \vspace{-5pt}
% \begin{minipage}{0.33\textwidth}
% \centering

% \input{figure_tex/1K0U_known}
% \end{minipage}
% \begin{minipage}{0.33\textwidth}
% \centering

% \input{figure_tex/1K1U_known}
% \end{minipage}
% \begin{minipage}{0.33\textwidth}
% \centering
% \input{figure_tex/2K1U_known}
% \end{minipage}
\caption{\textbf{Object Detection Performance of VLMs on Known Objects} in the \subref{fig:1K0U} 1K-0U (1 known and 0 unknown object), \subref{fig:1K1U} 1K-1U (1 known and 1 unknown objects), and \subref{fig:2K1U} 2K-1U (2 known and 1 unknown objects. Each baseline is run three times and performance's standard deviation is shown as vertical bar at each data point.}
\vspace{-10pt}
\label{fig:combine_detection}

\end{figure*}
We focus on conditional metrics given correct known-object classification: $p(x_{n \rightarrow n} \vert x_{k \rightarrow k})$: correct novel label assignment; and $p(x_{n \rightarrow k} \vert x_{k \rightarrow k})$: incorrect novel label assignment to a known object, introducing a normalized ME score:
% An effective model should maximize $p(x_{n \rightarrow n} \vert x_{k \rightarrow k})$ while minimizing $p(x_{n \rightarrow k} \vert x_{k \rightarrow k})$. We capture this trade-off using a normalized ME score:

% For this subtask, we first define $p(x_{n\rightarrow n})$ as the probability of correctly assigning the novel label to a novel object. For each IoU threshold $t$, $p(x_{n\rightarrow n})$ corresponds to the $AP@t$ metric for novel objects. This property is further equivalent to the ME score introduced in~\cite{gandhi2020mutual}. Similarly, we define $p(x_{k\rightarrow k})$ as the probability of correctly assigning the known label to a known object, and $p(x_{n\rightarrow k})$ as the probability of incorrectly assigning the novel label to a known object.

% To quantify ME bias, we focus on two key metrics: $p(x_{n\rightarrow n}\vert x_{k\rightarrow k})$ and $p(x_{n\rightarrow k}\vert x_{k\rightarrow k})$. Specifically, given that the known object is correctly identified, these metrics measure (1) the probability that the model also correctly detects the novel object in the same input image, and (2) the probability that the model mistakenly assigns the novel label to the known object. These metrics directly assess the model’s ability to apply ME bias when classifying novel objects. Ideally, we aim for a \underline{\textbf{high}} $p(x_{n\rightarrow n}\vert x_{k\rightarrow k})$ while keeping $p(x_{n\rightarrow k}\vert x_{k\rightarrow k})$ low. That means we want to maximize the normalized term:
\[
-1\leq\text{ME} = \frac{p(x_{n\rightarrow n}\vert x_{k\rightarrow k})-p(x_{n\rightarrow k}\vert x_{k\rightarrow k})}{p(x_{n\rightarrow n}\vert x_{k\rightarrow k})+p(x_{n\rightarrow k}\vert x_{k\rightarrow k})} \leq 1
\]
% We focus on the posterior term $x_{k \rightarrow k}$, as correct classification of known objects should lead a model with ME bias to assign the novel label to the novel object. 
% This enables evaluation of ME-based reasoning. 
The ME score is defined as: (1) $-1 \leq ME \leq 0$: Tendency to misassign the novel label to a known object, which indicates weak ME bias; (2) $0 < ME \leq 1$: Tendency to assign the novel label correctly. Higher ME score indicates stronger ME bias. In addition, the model may:
(1) Refuse to predict on novel labels: $p(x_{n \rightarrow \varnothing} \mid x_{k \rightarrow k})$, indicating conservative behavior;
(2) Misclassify background objects as novel or fail to accurately segment the objects, leading to low IoU: $p(x_{n \rightarrow bg} \mid x_{k \rightarrow k})$, often due to poor object detection.

\noindent\textbf{Spatial Reasoning.} 
% When multiple novel objects are present in the scene, the novel label may be incorrectly assigned to the wrong novel object. For example, the label ``dax" might be mistakenly assigned to the object ``blicket." 
To evaluate the model's spatial reasoning ability, we analyze the improvement in the probability of correctly assigning the novel label to the intended novel object, alongside the reduction in misclassifications among novel objects.
% A significant improvement in these metrics indicates that the model effectively leverages spatial information to disambiguate between novel objects. 
We quantify the impact of scene spatial descriptions on model performance by measuring the proportional increase in correctly assigning novel labels to novel objects when spatial context is provided compared to when it is absent. 
% This analysis highlights the extent to which each model benefits from additional spatial reasoning cues in improving novel object identification. 
Specifically,
\[
\text{Spatial Reasoning}=\frac{p_w(x_{n\rightarrow n}\vert x_{k\rightarrow k})-p_{w/o}(x_{n\rightarrow n}\vert x_{k\rightarrow k})}{p_{w/o}(x_{n\rightarrow n}\vert x_{k\rightarrow k})}
\]
where $p_w(.)$  and $p_{w/o}(.)$ represent the performance with and without spatial input, respectively. 
A positive Spatial Reasoning score indicates that the model effectively uses spatial context to enhance its performance, demonstrating a strong spatial reasoning capability.

When the scene contains multiple unknown objects, we additionally evaluate the ambiguity score, which quantifies the error rate of incorrectly assigning the novel label to the wrong novel object. This metric provides insight into the model's ability to disambiguate between multiple novel objects using available contextual cues.
\[
\text{Ambiguity}=\frac{p(x_{n\rightarrow no}\vert x_{k\rightarrow k})}{p(x_{n\rightarrow {n}}\vert x_{k\rightarrow k}) + p(x_{n\rightarrow no}\vert x_{k\rightarrow k})}
\]
where $p(x_{n\rightarrow {no}}\vert x_{k\rightarrow k})$ represents the probability of misassigning the novel label to the incorrect novel object, given that the known objects have been correctly identified. 
% Ideally, we want to minimize ambiguity score. 
% Higher ambiguity score indicates that the model is prone to aligning the novel name to the incorrect novel object. 

These metrics generalize to scenes with more objects. In multi-novel-object settings, $p(x_{n \rightarrow n} \mid x_{k \rightarrow k})$ captures both correct and incorrect assignments among novel objects, enabling broader evaluation of a model’s ability to manage complexity and ambiguity.

% Note that these metrics can be generalized beyond the current settings to accommodate scenes with a greater number of objects. For instance, in scenarios where multiple novel objects are present, the term $p(x_{n\rightarrow {n}}\vert x_{k\rightarrow k})$ in the ME score accounts for both cases: when the model correctly assigns the novel label to the intended novel object, as well as when it misassigns the novel label to the incorrect novel object. This extension allows for a more comprehensive evaluation of a model’s ability to handle increased scene complexity and ambiguity in object identification.

% \noindent\textbf{Implementation Details.} We evaluate all baselines three times across all settings to mitigate the influence of random factors that could impact performance, such as viewpoint variations, lighting conditions, and scene composition. 
% This ensures a more reliable and robust assessment of each model's capabilities.

\section{Results}
\subsection{Object Detection}

\noindent\textbf{High Performance on Known Objects.} In Figure~\ref{fig:combine_detection}, we show the performance of various baselines in the 1K-0U setting. Except for LLaVA-OV—primarily designed for general VQA and reasoning rather than object detection or grounding—most methods achieve consistently high performance. When an unknown object is introduced (Figure~\ref{fig:1K1U}), performance shows only a slight decline across all methods, indicating strong robustness in detecting common, known objects.
% Notably, CogVLM~\cite{wang2023cogvlm} experiences a rapid decline in performance at higher IoU thresholds, suggesting that while its predicted bounding boxes are correctly localized, they exhibit slight misalignments relative to the ground truth. In contrast, most other methods are explicitly trained to produce segmentation masks instead of bounding boxes, leading to more precise predictions. Gemini~\cite{geminiteam2024geminifamilyhighlycapable} also shows a steeper performance drop at higher IoU thresholds, further indicating potential localization inconsistencies.

% \noindent\textbf{Background Distractions Have Low Impact on Performance.} In~\ref{fig:1K_detection}, we observe that the presence of background distractions has a minimal impact on model performance, with only a slight decline at higher IoU thresholds. This suggests that these vision-language models (VLMs) exhibit strong robustness in object detection, effectively distinguishing target objects from irrelevant elements and avoiding confusion caused by background patterns.

\noindent\textbf{Extra Known Object Decreases Performance.} In Figure~\ref{fig:2K1U}, we observe a significant performance drop across most models when an additional known object is introduced into the scene. This decline can be attributed to increased clutter, occlusion, and potential confusion between similar known objects. However, CogVLM~\cite{wang2023cogvlm} demonstrates greater robustness in object detection across both settings, effectively distinguishing objects despite the added complexity.

\subsection{VLMs Show Weak ME Bias}

In this subsection, we examine the ME bias of baseline models by analyzing the distribution of their responses to novel labels, specifically in the case where they have correctly identified the known object(s). This analysis provides insights into how effectively each model applies the ME assumption when associating novel labels with novel objects. In these settings we set the threshold $t$ in the $AP@t$ metric to $0.5$ for evaluation.

\begin{figure*}
\begin{minipage}[c]{0.33\textwidth}
\centering
	\includegraphics[width=0.9\linewidth]{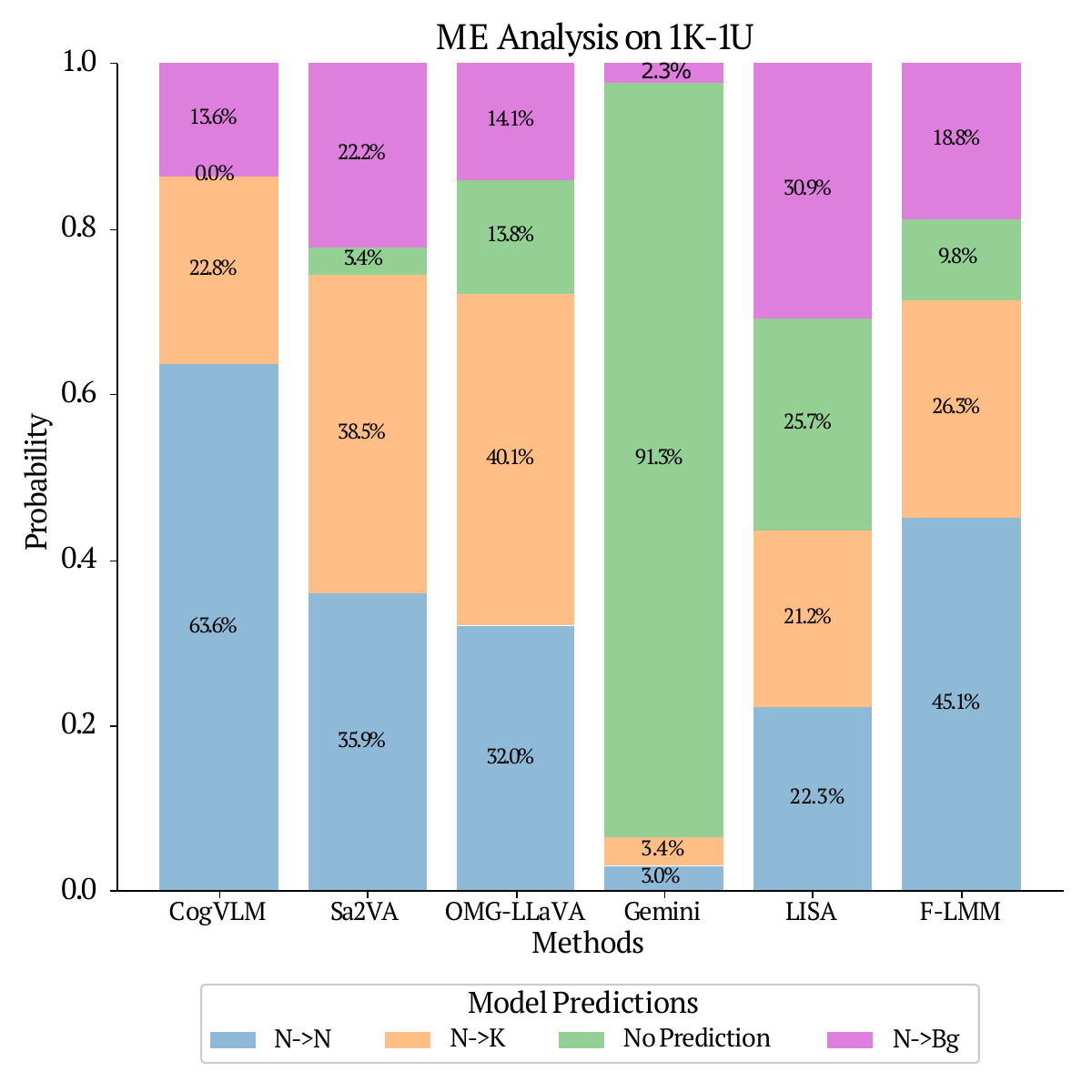}
	% \caption{\textbf{Non-overlapping 2D Mask Proposal.} We address the issue of overlapping masks produced by SAM. The masks are first sorted by their areas. Subsequently, the smaller masks are stacked on top of the larger ones. Non-overlapping masks are obtained by taking the visible segment of each mask.}

	% \label{fig:1K_1U_fig}
  % \vspace{-5pt}
   % \caption{\label{fig:1K_1U_fig}}
\end{minipage}
\begin{minipage}[c]{0.33\textwidth}
\centering
	\includegraphics[width=0.9\linewidth]{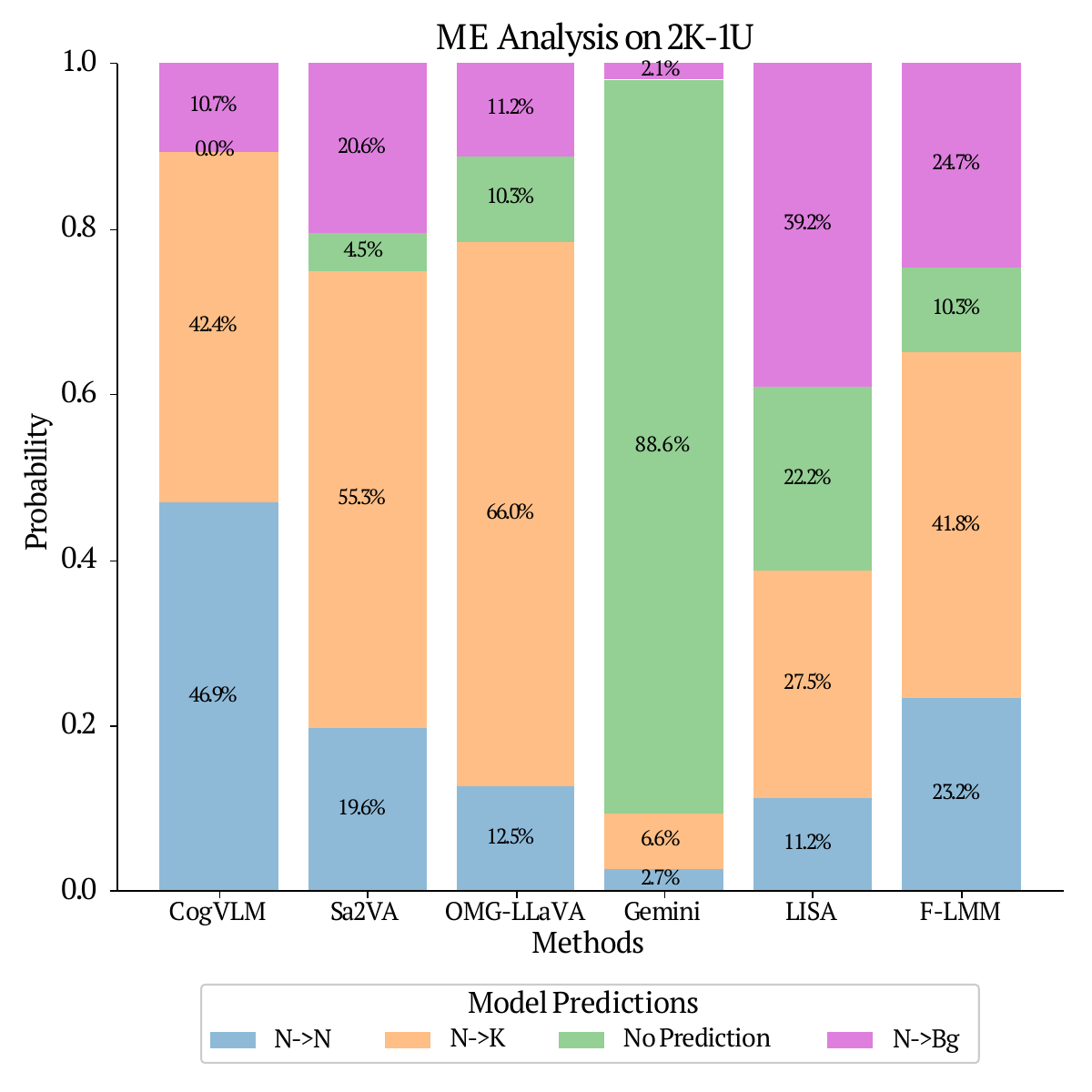}
	% \caption{\textbf{Non-overlapping 2D Mask Proposal.} We address the issue of overlapping masks produced by SAM. The masks are first sorted by their areas. Subsequently, the smaller masks are stacked on top of the larger ones. Non-overlapping masks are obtained by taking the visible segment of each mask.}

	% \label{fig:2K_1U_fig}
  % \vspace{-5pt}
   % \caption{\label{fig:2K_1U_fig}}
\end{minipage}
\begin{minipage}[c]{0.33\textwidth}
\centering
	% \includegraphics[width=0.8\linewidth]{figures/2K-1U-barchart.pdf}
% \centering
	\begin{center}
% \caption{PartNetE}

\begingroup
\setlength{\tabcolsep}{3pt} % Default value: 6pt
\renewcommand{\arraystretch}{1.} % Default value: 1

% \vspace{-10pt}
\scalebox{0.9}{

% \begin{table}[h]
% \centering
 \begin{tabular}{l|c  c}
% \caption{PartNetE}
 
        \toprule
        Method & 1K-1U $\uparrow$ & 2K-1U $\uparrow$\\
        \midrule
        CogVLM & 0.472 & 0.050\\
        F-LMM & 0.263 & -0.286\\
        LISA & 0.025 & -0.421\\
        Sa2VA &  -0.035 & -0.477\\
        Gemini & -0.063 & -0.419\\
        OMG-LLaVA & -0.112 & -0.682\\

        \bottomrule

\end{tabular}
% \end{table}

}
% \caption{.}
\endgroup
 % \label{tbl:me_score_2K1U}

\end{center}

\end{minipage}

\caption{\textbf{Mutual Exclusivity (ME) Analysis in the (Left) 1K-1U and (Middle) 2K-1U settings}. These settings contain one novel object in the scene. The response types are categorized as follows: $N\rightarrow N$ denotes correctly assigning the novel label to the novel object, $N\rightarrow K$  represents misassigning the novel label to a known object, and $N\rightarrow Bg$ indicates misassigning the novel label to a background distractor or failing to detect high-quality object bounding boxes. Additionally, No Prediction indicates cases where the model fails to produce a bounding box for the referred object. \textbf{(Right) ME Scores} of 1K-1U and 2K-1U settings. Higher scores indicate stronger ME bias.} \label{fig:me_combined_1k}
\vspace{-10pt}
\end{figure*}

In Figures~\ref{fig:me_combined_1k} (left and middle), we present the response distributions of six baseline models, along with their ME scores (Figure~\ref{fig:me_combined_1k}, right), computed using the formula introduced in Section~\ref{sec:chap6_metric}. Among all baselines, CogVLM~\cite{wang2023cogvlm} achieves the highest ME score, indicating a strong mutual exclusivity bias. F-LMM~\cite{wu2024f} and LISA~\cite{lai2024lisa} follow, though with a significant performance gap. The remaining methods exhibit negative ME scores, suggesting a weaker ME bias, as they frequently misassign novel labels to known objects, failing to leverage ME. Introducing an additional known object leads to a notable drop in the ME scores. Specifically, we observe a significant decrease in correctly assigning the novel label to the novel object, and a corresponding increase in misassigning the novel label to known objects, as reflected in the blue and orange segments of the bar charts. This suggests that introducing an additional known object into the scene substantially increases task difficulty, leading to a significant decline in ME scores ($\Delta_{avg}=0.464$). Additionally, the scene becomes more cluttered and occluded, further challenging the model's ability to accurately distinguish between known and novel objects.
% With two known objects present, the model faces twice the risk of misassigning the novel label, leading to a greater decline in ME scores. Additionally, the scene becomes more cluttered and occluded, further challenging the model's ability to accurately distinguish between known and novel objects.

% In~\ref{fig:me_combined_2k}, we observe a similar trend in the ME scores across models; however, all scores are significantly lower compared to~\ref{fig:me_combined_1k}. This suggests that introducing an additional known object into the scene substantially increases task difficulty. With two known objects present, the model faces twice the risk of misassigning the novel label, leading to a greater decline in ME scores. Additionally, the scene becomes more cluttered and occluded, further challenging the model's ability to accurately distinguish between known and novel objects.

% \input{tables_tex/chapter6/combine_me_spatial_2k}

% CogVLM~\cite{wang2023cogvlm} is the only method that exhibits a positive ME score in this setting, demonstrating a stronger mutual exclusivity bias. In contrast, all other models show a tendency to misassign novel labels to known objects, even when they have correctly identified the known objects in the scene. This suggests that while these models can recognize known objects, they struggle to apply mutual exclusivity reasoning to correctly associate novel labels with novel objects.

Notably, Gemini~\cite{geminiteam2024geminifamilyhighlycapable} has the highest rate of missing predictions for novel objects, failing to produce a bounding box nearly 90\% of the time. This behavior can be attributed to its training paradigm, which prioritizes reducing hallucinations in general conversations. In contrast, CogVLM~\cite{wang2023cogvlm} consistently produces a prediction, with a 0\% no-prediction rate, as it is trained on grounding datasets that enforce object localization for every referred object. As a result, CogVLM~\cite{wang2023cogvlm} does not fail to detect objects but may still struggle with accurate label assignment.
Additionally, LISA~\cite{lai2024lisa} exhibits the highest value of $p(x_{n\rightarrow bg}\vert x_{k\rightarrow k})$, suggesting weaker object localization capabilities compared to other models.

% In summary, the results indicate that CogVLM~\cite{wang2023cogvlm} exhibits the strongest ME bias, while OMG-LLaVA~\cite{zhang2025omg} demonstrates the weakest. Other models show a higher tendency to misassign the novel label to known objects, suggesting a weaker ME bias and a reduced ability to leverage the ME assumption for novel object detection.

% \input{figure_tex/chapter6/combined_2u}

\subsection{VLMs Leverage Spatial Context To Reduce Ambiguity}
We examine the models' ability to leverage spatial context to disambiguate between multiple novel objects when more than one is present in the scene. 
% Additionally, we analyze whether extra spatial context helps correct errors made by the models in cases where no inherent ambiguity exists, such as in the 2K-1U-YesD setting. This allows us to assess whether scene context serves as a critical factor in improving model accuracy, even when the task does not explicitly require disambiguation.
% With the inclusion of spatial input, the performance of all models significantly improves, as reflected in the increase in ME scores (see~\ref{fig:me_combined_2k_spatial}). The spatial reasoning metric quantifies the extent to which additional scene context helps models refine their predictions. Among all methods, Gemini~\cite{geminiteam2024geminifamilyhighlycapable} benefits the most from the scene description input, achieving the highest spatial reasoning score, indicating strong utilization of spatial context. In contrast, while CogVLM~\cite{wang2023cogvlm} shows some improvement with spatial input, it does not effectively leverage this information, as evidenced by its lowest spatial reasoning score.
Table~\ref{tbl:combined_score_2U}  illustrates model performance in the 1K-2U setting, where two novel objects are present in the scene. When scene context is absent, ambiguity arises between the two novel objects, leading most models to exhibit a high ambiguity score which is an expected outcome, as there are no additional cues available for distinguishing between the two novel objects. However, when scene context is provided, the ambiguity score drops significantly across all models, demonstrating that models can effectively leverage spatial information to resolve ambiguity and correctly assign novel labels.

\begin{table*}[t]
\caption{\textbf{Impact of Spatial Scene Input on Ambiguity \& Spatial Reasoning Scores.} We compare settings with and without spatial scene input, evaluating models based on two key metrics: Ambiguity Score, which measures the ability to disambiguate novel objects (lower is better), and Spatial Reasoning Score, which reflects how effectively models utilize spatial context to improve performance, with respect to the setting without spatial context (higher is better). Bold values indicate the most desirable outcomes.}
\begin{center}

\begingroup
\setlength{\tabcolsep}{3pt} % Default value: 6pt
\renewcommand{\arraystretch}{1.} % Default value: 1

% \vspace{-10pt}
\scalebox{0.9}{

% \begin{table*}[h]
% \centering
% \caption{PartNetE}
 \begin{tabular}{l|c c c c c c}
        \toprule
        Method & CogVLM & Sa2VA & OMG-LLaVA & Gemini & LISA & F-LMM \\
        \midrule
        Ambiguity without scene dec. $\downarrow$ & 0.525 & 0.480 & 0.500 & 0.510 & 0.394 & 0.510 \\
        Ambiguity with scene dec. $\downarrow$ & 0.384 & \textbf{0.258} & 0.345 & \textbf{0.257} & 0.283 & 0.365\\
        Spatial Reasoning (Relative to w/o scene dec.) $\uparrow$& 0.314 & 0.914 & 1.012 & \textbf{11.167} & 1.853 & 1.314 \\
        Spatial Reasoning (Absolute) $\uparrow$& 0.117 & 0.203 & 0.165 & \textbf{0.268} & 0.265 & 0.230 \\
        
       \bottomrule

\end{tabular}

}

\endgroup

\end{center}

 \label{tbl:combined_score_2U}
 % \vspace{-10pt}

\end{table*}

% Similar to the 2K-1U-YesD setting, 
Gemini~\cite{geminiteam2024geminifamilyhighlycapable} benefits the most from the additional spatial input, achieving the highest spatial reasoning score, indicating strong contextual refinement in label assignment. Other models also show improvements in spatial reasoning, though to varying degrees. CogVLM~\cite{wang2023cogvlm}, once again, ranks the lowest in its ability to effectively utilize spatial context for reasoning.

% In summary, while CogVLM~\cite{wang2023cogvlm} shows some improvement when provided with spatial scene information, it fails to effectively utilize this context compared to other models. In contrast, Gemini~\cite{geminiteam2024geminifamilyhighlycapable} demonstrates a strong ability to leverage spatial input, significantly refining its predictions and improving overall performance.

\subsection{Understanding The Use of Object Names in the ME Task}
In this subsection, we analyze additional factors that may influence model performance, beyond solely ME bias and spatial reasoning, using the 2K-1U setting. 
We select this setting because it presents a greater visual challenge compared to 1K-1U, providing a larger margin for improvement and deeper insights into model behavior. Additionally, this setting is inherently deterministic: a model leveraging ME bias can solve the task without relying on extra contextual information from the scene. This allows us to systematically evaluate the impact of additional information and determine whether it contributes to performance improvements.

%%%% ABC, Toys, ShapeNet datasets
\begin{figure}[t]
\centering
\includegraphics[width=0.9\linewidth]{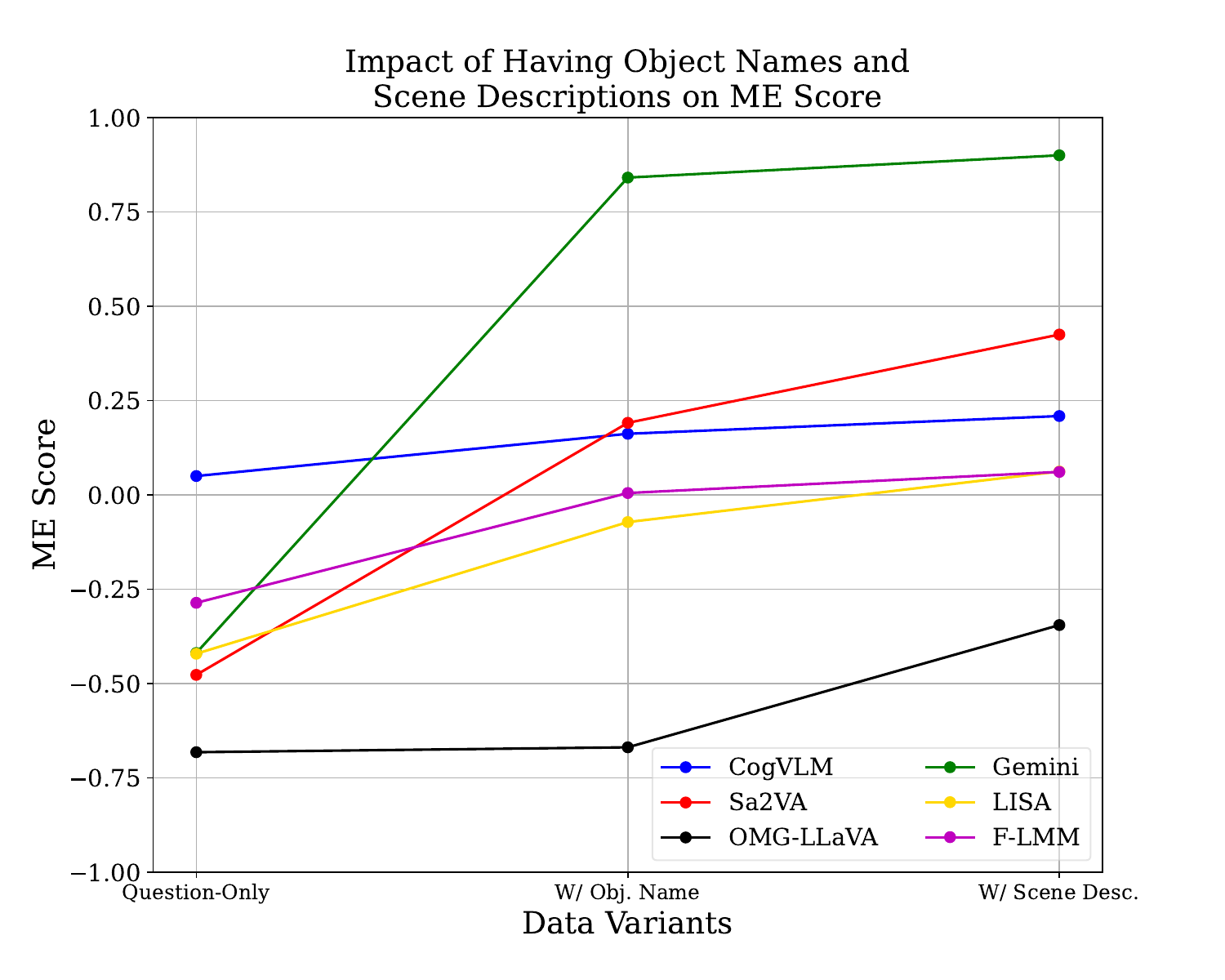}
\caption{\textbf{Impact of Having Object Names and Scene Descriptions on ME Score} comparing between Question-Only language prompt (e.g. Where is the dax?); Minimal Scene Context (e.g. There are three objects in the scene: dog, a cat, and a dax. Where is the dax?) and Full Scene Description: The models receive a detailed scene description}
\label{fig:me_2k1u_3variants}
\vspace{-15pt}
\end{figure}
% \caption{}
% \label{fig:setting_overview}
% \vspace{-20pt}

% \noindent\textbf{Impact of Providing Object Names.} 
In Figure~\ref{fig:me_2k1u_3variants}, we present the models' performance across three different prompt settings: 
(1) Question-Only Prompt: The prompt consists solely of the query, e.g., ``Where is the dax?"; (2) Minimal Scene Context: The prompt includes only the object names present in the scene, e.g., "There are three objects in the scene: a dog, a cat, and a dax. Where is the dax?"; (3) Full Scene Description: The models receive a detailed scene description as outlined in Section~\ref{sec:data_variants}.

By comparing performance between (1) Question-Only and (2) Minimal Scene Context, we assess the impact of knowing the object names present in the scene. The difference between (2) Minimal Scene Context and (3) Full Scene Description further evaluates the models' ability to leverage spatial reasoning when additional context is provided.

All models show performance improvements when provided with Minimal Scene Context (i.e., object names).
% with Sa2VA~\cite{yuan2025sa2va} and Gemini~\cite{geminiteam2024geminifamilyhighlycapable} exhibiting the most significant gains in ME score.
% This increase highlights the advantage of even minimal contextual information over the Question-Only setting.
This result aligns with expectations: while the additional information consists solely of object labels, it implicitly signals to the model that ``dax" is distinct from ``dog" and ``cat", reinforcing the mutual exclusivity assumption and aiding in accurate object-label associations. 
% Furthermore, having access to object names reduces the likelihood of models incorrectly assigning the novel label to background distractions, improving their ability to distinguish between foreground objects and irrelevant scene elements.

% \input{figure_tex/chapter6/masks_object_given}

With the exception of CogVLM~\cite{wang2023cogvlm}, which performs similarly to the Minimal Scene Context setting, all other models show improved performance in the Full Scene Description case, where spatial information is provided. This demonstrates the benefit of incorporating spatial context in aiding object reasoning and disambiguation. 
\section{Limitations}
Multiple additional factors influence mutual exclusivity bias in real-world settings, including social interaction and temporal ambiguity, which we leave to future work. Although our setup captures uncertainty reasoning, scene-level spatial descriptions represent only one possible way to resolve ambiguity. In practice, children and adults rely on many other mechanisms, such as accumulated experience over time and broader reasoning strategies. Future work could investigate alternative ways of resolving ambiguity under uncertainty. As with all studies on LLMs and VLMs, performance can also be highly sensitive to prompt design, and different models may respond best to different prompts. In this work, we use the prompts recommended by the original authors for each method in object detection and segmentation settings. Future research could explore alternative prompting strategies both to better understand model behavior and to further improve their reasoning performance.

\section{Conclusion}
We introduce MEBench, a novel benchmark designed to study mutual exclusivity (ME) bias and to extend this analysis toward spatial reasoning in object-label mapping tasks for SOTA VLMs. We assess several SOTA VLMs on MEBench using a scalable synthetic data generation pipeline. These models exhibit weak ME bias, and their performance declines sharply even under modest increases in scene complexity. Furthermore, providing additional scene context helps reduce ambiguity and improves performance. However, these models remain far from an ideal solution, one that would seamlessly integrate strong object localization, ME bias, and spatial reasoning. Bridging this gap remains a crucial direction for future research, as these properties are essential for developing more advanced, human-like reasoning in AI models. We hope that MEBench will encourage the AI community to systematically explore the interplay between ME bias, spatial reasoning, and multimodal learning, and to develop models with stronger cognitive capabilities.
\clearpage
\appendix
\begin{center}
{\scshape\LARGE Appendix \par}
\end{center}

% This supplemental material is structured as follows: We first provide more details about data information in Section~\ref{appdx:data}. We then show additional results in Section~\ref{appdx:exp}. Finally, we provide additional training details about our baseline models in Section~\ref{appdx:models}.

\section{Data}\label{appdx:data}

\subsection{Datasets}
In this work, we use Toys4K~\cite{stojanov2021using} as the primary dataset for known objects in our experiments. Toys4K consists of 4,179 object instances spanning 105 categories, collected under Creative Commons and royalty-free licenses. This diverse and openly licensed dataset provides a rich foundation for evaluating model performance on familiar object categories.

\noindent\textbf{Novel Objects.} We introduce a curated set of 64 novel objects, sourced from GeoShapeV2~\cite{geoshapes} and Thingi10K~\cite{zhou2016thingi10k}. These objects are manually designed and procedurally generated using geometric nodes in Blender~\cite{blender} with diverse structures. During rendering, these novel objects are assigned randomized materials and colors, ensuring diversity in appearance and preventing models from relying on any texture-based shortcuts for recognition.

\subsection{Data Generation Pipeline}
\begin{figure*}[t!]
\centering
\includegraphics[width=\linewidth]{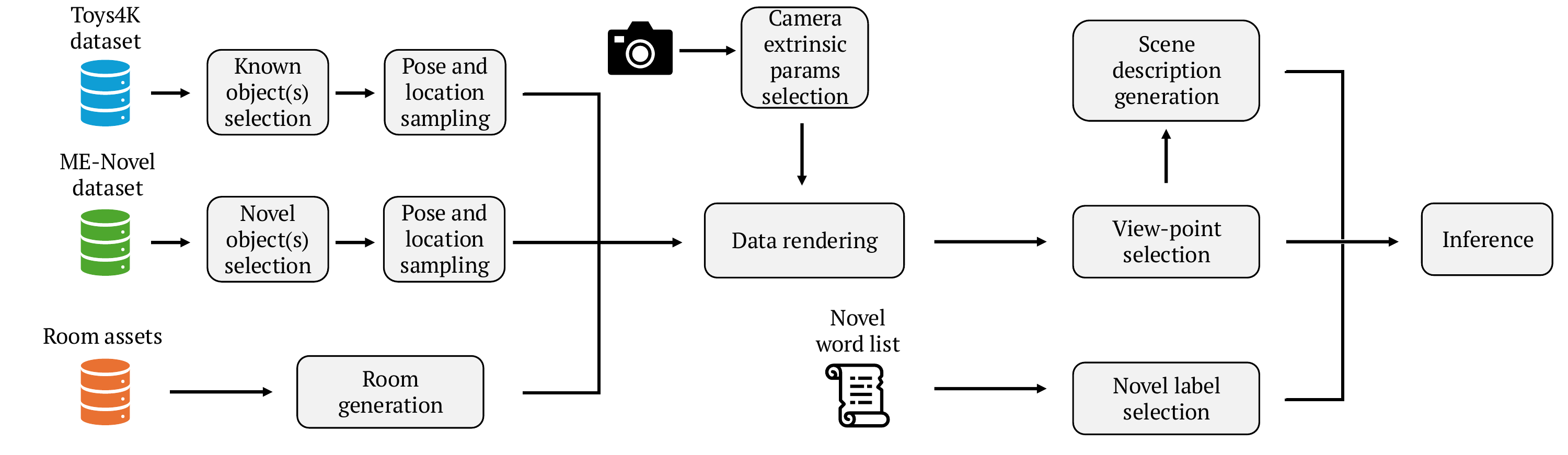}
\caption{\textbf{Data Generation Pipeline.}  We begin with 3D databases containing known objects, novel objects, and background room assets, from which we select and compose components into a 3D scene. The scene is then rendered from multiple camera viewpoints. During inference, we select a viewpoint where all objects are visible. Based on this selected view, we generate a spatial scene description and assign a novel label to each novel object, drawing from a list of randomly generated words.}
\label{fig:data_generation_pipeline}
% \vspace{-20pt}
\end{figure*}

In this subsection, we describe our data generation pipeline. An overview can be seen in Figure~\ref{fig:data_generation_pipeline}. Our system is designed to be compatible with any 3D categorical dataset. 
% For our benchmark, we use the Toys4K~\cite{stojanov2021using} dataset, chosen for its large number of categories (105) and diverse toy-like object appearances, which closely resemble the variability encountered in real-world child learning scenarios. This dataset comprises common, everyday object categories (e.g., "car", "dog"), which serve as the known categories in our benchmark.

%%%% ABC, Toys, ShapeNet datasets
\begin{figure*}[h]
\centering
\includegraphics[width=0.8\linewidth]{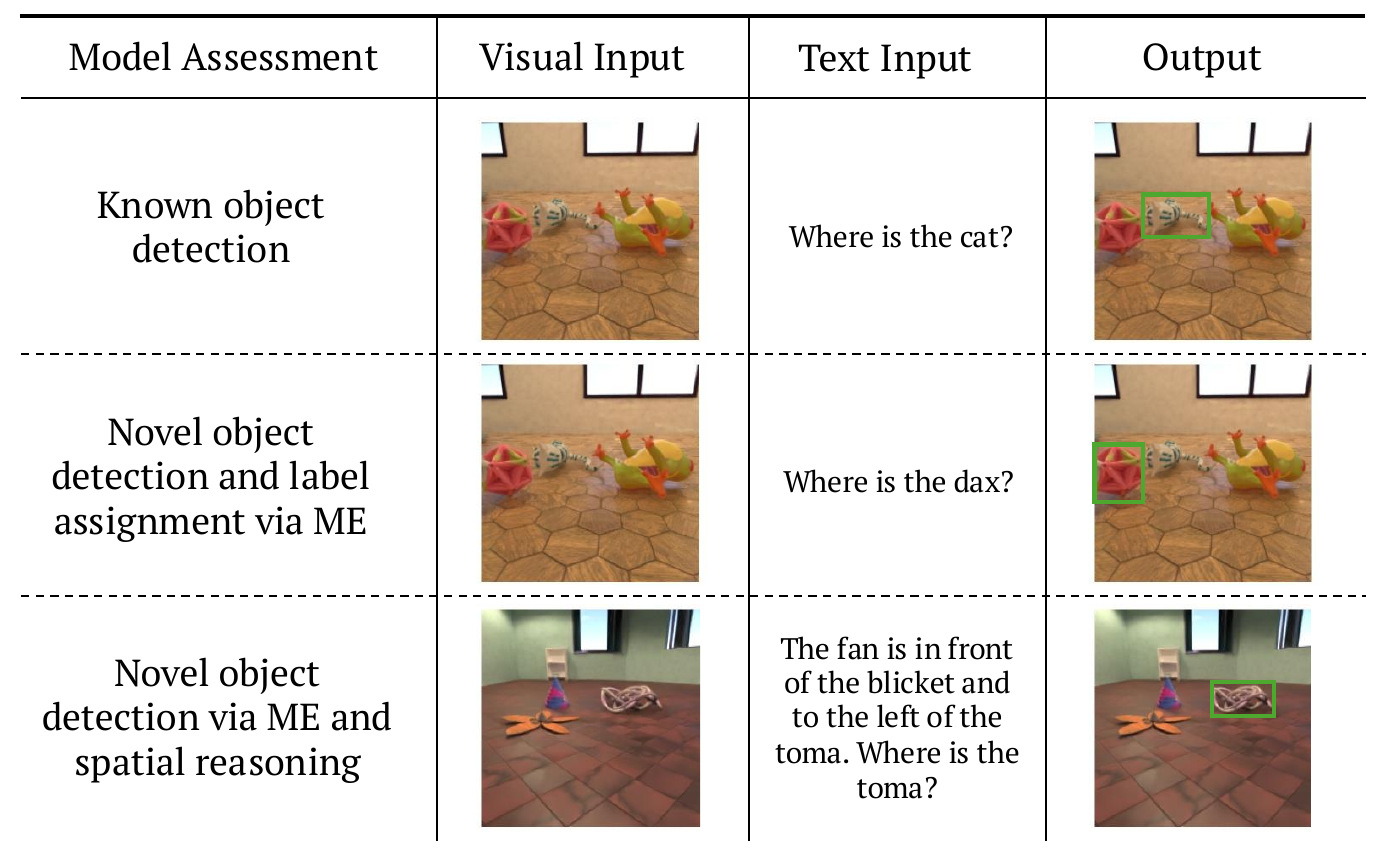}
\caption{\textbf{Illustrative Example of Expected Inputs and Outputs in MEBench.} For
each subtask, we present the expected visual input, text input, and the corresponding model
output, demonstrating the structured evaluation process.}
\label{fig:input_output}
% \vspace{-15pt}
\end{figure*}
% \caption{}
% \label{fig:setting_overview}
% \vspace{-20pt}

\noindent\textbf{Background.} To create diverse and realistic backgrounds, we generate room environments as the backdrop for our scenes. These rooms are primarily living rooms and bedrooms, as they represent the most natural settings for child play. Our room generation is based on Infinigen~\cite{infinigen2024indoors}, ensuring procedural diversity. Each room includes varied background object configurations, such as TVs, plants, shelves, beds, and other household items. Lighting conditions are naturally derived from indoor sources (e.g., lamps, ceiling lights) or outdoor light filtering through open doors and windows.

\noindent\textbf{Data Rendering.} During each scene rendering, we first randomly select a subset of objects from the known categories and a set of novel objects from our curated collection. The total number of objects and the number of novel instances are determined based on user-defined input parameters. To achieve realistic object placement, we use rigid body simulation to generate natural rotational poses. Objects are then scaled and positioned at random locations within the scene while ensuring that no two objects collide. Further, we prevent intersections between placed objects and background elements (e.g., plants, furniture, walls). To enhance diversity and naturalistic scene appearance, we sample multiple camera viewpoints for each scene, capturing variations in perspective, depth, and occlusions.

\noindent\textbf{Scene Description Generation.} We generate scene descriptions as additional contextual inputs for each view of the scene, rather than for the entire 3D scene (see Figure~\ref{fig:data_generation_pipeline}). This is because some spatial relationships in 3D are inherently viewpoint-dependent terms like ``left," ``right," ``in front of," and ``behind" can vary depending on the observer’s perspective.

For each view, we first construct a scene graph using object bounding boxes and the corresponding depth map. In this graph, objects are represented as nodes, while pairwise spatial relationships form the edges. We then translate the structured scene graph representation into plain English descriptions. For instance, given the scene graph expression: ``Dog":\{``to left of":[``dax", ``pig"]\} we generate the natural language description: ``The dog is to the left of the dax and the pig." Notably, this scene description generation process is fully deterministic, ensuring that all pairwise object relationships are consistently included in the description. This eliminates potential ambiguities and provides a structured yet flexible input format for downstream reasoning tasks.

We generate the 3D scene before creating scene descriptions for each view because spatial relationships in 3D can be ambiguous. For example, directional terms such as behind, in front of, left, and right depend on the viewpoint and cannot be directly described in 3D space but can be precisely defined in 2D images. Therefore, we first generate the 3D scene before generating scene graph and translate it into English descriptions for each specific view.

\noindent\textbf{View-point Selection for Inference.} During inference, we select rendered viewpoints where all objects in the scene are visible. An object is considered visible in a given view if its segmentation mask occupies at least 200 pixels in a $224\times 224$ resolution image. This criterion ensures that: 1) All models are evaluated fairly, eliminating potential viewpoint bias where certain views contain fewer visible objects than others; 2) The designated experimental setting is strictly maintained, for instance, in the 2K-1U setting, all three objects (two known, one unknown) are always visible during inference. To further minimize variability in object visibility, we run each model three times, each with a different set of selected viewpoints.

%%%% ABC, Toys, ShapeNet datasets
% \caption{}
% \label{fig:setting_overview}
% \vspace{-20pt}

\subsection{Licensing \& Open-source}
All assets used in this benchmark were collected under Creative Commons or Royalty-Free licenses. The object-level attributes for Toys4K are available on the project’s GitHub repository. All objects from GeoShapeV2 are licensed as Royalty-Free. Additionally, we selected six objects from Thingi10K: 100423, 42370, 43664, 59771, 93073, and 94146.

The room background generation in our data pipeline is adapted from Infinigen~\cite{infinigen2024indoors}, which is released under the BSD 3-Clause License.

We release the generated data for all experimental settings, along with the open-source data generation code at \url{https://github.com/ngailapdi/MEBench}. Additionally, we provide inference and evaluation code for all VLM baselines to facilitate reproducibility and future research.

\section{Baselines}
% \begin{table}[t!]
% \renewcommand{\arraystretch}{1.0}
% \begin{center}
% \caption{Performance of DINOv2 and our method fine-tuned on Toys4k and ABC on Toys4k under LSME setting. All methods use ViT B/14 as the backbone and our method is initialized with pretrained DINOv2 weights. Training on ABC improves the performance significantly, surpassing the model that was trained on the base classes of Toys4k with the same number of scenes.}
% \scalebox{0.9}{
% \begin{tabular}{l|c|c}\\
% Method & LSA & SA \\
% \hline

% DINOv2 & 39.24\ci{1.17}&50.88\ci{1.91}\\
% Ours-DINOv2-Toys & 43.62\ci{1.29}&53.44\ci{1.89}\\
% Ours-DINOv2-ABC &\best{47.70\ci{1.26}}&\best{61.32\ci{1.86}}

% \label{tbl:multi-finetune}
% \end{tabular}}
% \end{center}
% \vspace{-20pt}
% \end{table}

\begin{table*}[t]
\renewcommand{\arraystretch}{1.1}
\begin{center}
\caption{\textbf{Overview of Vision-Language Model (VLM) Baselines.} We categorize evaluated models based on their output types: (1) Text-only, (2) Text + Bounding Box, and (3) Text + Segmentation Mask. Each model's venue and primary focused tasks are listed.}
\scalebox{0.7}{
\begin{tabular}{c|l|c|l}
\toprule
\textbf{Model Output Type} & \textbf{Method} & \textbf{Venue} & \textbf{Focused Tasks} \\
\midrule
\multirow{1}{*}{Text-only} 
& LLaVA-OV & arXiv 2024 & Open-sourced general VQA and reasoning \\
\midrule
\multirow{3}{*}{\shortstack{Text + \\ Bounding Box}} 
& CogVLM-Grounding & NeurIPS 2024 & Referring expression comprehension, grounding VQA \\
& Gemini 2.0 Flash & \begin{tabular}{@{}c@{}}Google DeepMind \\ Blog Post 2024 \end{tabular} & Closed-source general VQA and grounding \\
\midrule
\multirow{5}{*}{\shortstack{Text + \\ Segmentation Mask}} 
& LISA & CVPR 2024 & Reasoning-based segmentation \\
& OMG-LLaVA & CVPR 2024 &  Reasoning + Referring segmentation\\
& Sa2VA & arXiv 2025 & Referring segmentation\\
& F-LMM & arXiv 2024& Referring segmentation\\
\bottomrule
\end{tabular}}
\label{tbl:vlm_baseline}
\end{center}
\end{table*}

% DINOv2 & ViT S/14 & 41.26\ci{1.15} &52.28\ci{1.86} & 37.08\ci{1.05}&48.16\ci{1.87}\\
% Ours-DINOv2-Toys & ViT S/14 & 45.54\ci{1.27}&55.20\ci{1.96} &40.38\ci{1.15} &50.64\ci{1.87} \\
% Ours-DINOv2-ABC & ViT S/14 & 50.37\ci{1.25}&\sbest{64.64\ci{1.89}} &\sbest{43.92\ci{1.23}} &\sbest{57.88\ci{1.97}}\\
% \vspace{-10pt}\\
% \hline

% DINOv2 & ViT B/14 & 43.21\ci{1.21}& 54.96\ci{1.89}& 39.24\ci{1.17}&50.88\ci{1.91}\\
% Ours-DINOv2-Toys & ViT B/14& \sbest{51.89\ci{1.35}}&62.60\ci{1.80} & \sbest{43.62\ci{1.29}}&53.44\ci{1.89}\\
% Ours-DINOv2-ABC & ViT B/14& \best{55.09\ci{1.30}}& \best{68.40\ci{1.79}} &\best{47.70\ci{1.26}}&\best{61.32\ci{1.86}}
We evaluate our MEBench benchmark on seven SOTA VLM baselines, consisting of both closed-source and open-source models: CogVLM~\cite{wang2023cogvlm}, Gemini~\cite{geminiteam2024geminifamilyhighlycapable}, Sa2VA~\cite{yuan2025sa2va}, OMG-LLaVA~\cite{zhang2025omg}, LISA~\cite{lai2024lisa}, LLaVA-OV~\cite{li2024llava}, and F-LMM~\cite{wu2024f}. These baselines are categorized into three primary model types based on their output modalities: (1) Text-only output, (2) Text + Bounding Box output, and (3) Text + Segmentation Mask output. Each of these models has been pre-trained and fine-tuned on large-scale datasets, achieving SOTA performance across various vision-language tasks and benchmarks. For a comprehensive summary of these baselines, please refer to Table~\ref{tbl:vlm_baseline}.

The majority of these VLMs are trained primarily for object grounding—performing object detection or segmentation based on text prompts. In contrast, LLaVA-OV~\cite{li2024llava} is designed as a more general VQA and reasoning model. The diversity in model architectures and training objectives allows us to conduct a comprehensive analysis, gaining deeper insights into how different VLMs perform on our benchmark across various reasoning and perception challenges.

\section{Experiment Details}
\noindent\textbf{Experimental Compute Resources.} All baselines were evaluated using a single NVIDIA RTX 4090 GPU with 24GB of memory. CogVLM~\cite{wang2023cogvlm} required approximately 2 hours of wall-clock time for inference, while the remaining models completed inference in 1–2 minutes.

\noindent\textbf{Implementation Details.} We evaluate all baselines across all settings using three independent runs, and report the results as the mean of these runs. For each baseline, we adopt the optimal prompt as recommended by the original authors.
% to mitigate the influence of random factors that could impact performance, such as viewpoint variations, lighting conditions, and scene composition.

\section{Limitations}
\noindent\textbf{Benchmark Design.} There are multiple factors that affect ME bias in real life, including social interactions, temporal ambiguity and label structures. In this work we only model the core axes of the problem, that are the number of known and novel objects in the scene and whether background distractors exist, leaving others for future work. 

While our setting models uncertainty reasoning, providing scene spatial description is only one of the ways to resolve ambiguity. There are multiple other methods that children and adults use to resolve such ambiguity in real life, including accumulated priors over time and other reasoning strategies. Future work can explore more methods to resolve ambiguity in the case of uncertainty.

Due to the procedural nature of our data generation system, we have the ability to increase the task complexity, for instance, adding more known and novel objects or more cluttered background. Future research can explore more challenging settings and developing approaches that tackle highly complex scene configurations.

\noindent\textbf{Evaluation.} We assume that the models possess an understanding of ``objectness", meaning they can recognize and parse all objects present in the scene. This assumption is reasonable because these models are strong object detectors, trained on large-scale datasets covering a diverse range of objects. Additionally, the data is rendered to ensure that objects remain largely visible within the scene, making it visually easier for models to accurately parse and recognize all objects present. Consequently, the errors analyzed in this study do not account for failures in basic scene parsing but rather focus on higher-level reasoning and learning biases.

\noindent\textbf{Prompt Design.} As with all studies on LLMs and VLMs, model performance can be highly dependent on prompt design, with different models responding better to different prompts. In this work, we follow the prompts recommended by the original authors for each method in object detection and segmentation tasks. Future research can explore alternative prompt engineering strategies to gain deeper insights into model performance and optimize their reasoning capabilities.

{
    \small
    \bibliographystyle{ieeenat_fullname}
    \bibliography{main}
}


\end{document}